\documentclass[journal,onecolumn]{IEEEtran}
\usepackage{amsmath,amsfonts}
\usepackage{algorithmic}
\usepackage{array}
\usepackage[english]{babel}
\usepackage[utf8]{inputenc}
\usepackage{verbatim}
\usepackage{subfigure}
\usepackage{graphicx}
\usepackage{booktabs}

\begin{document}

%\markboth{Submitted to IEEE TNNLS
%{\em Special Issue on
%Causal Discovery and Causality-Inspired Machine Learning}}{SKM: My IEEE article}    

\title{Causal Effect Estimation using Variational Information Bottleneck}

%\author{Zhenyu Lu, Yurong Cheng, Mingjun Zhong, George Stoian, Ye Yuan, Guoren Wang\\
%\small School of Computer Science, Beijing Institute of Technology\\
%        \small Department of Computing Science, University of Aberdeen 
%}
%\date{}

\author{\IEEEauthorblockN{Zhenyu Lu\IEEEauthorrefmark{2},
Yurong Cheng\IEEEauthorrefmark{2}, Mingjun Zhong\IEEEauthorrefmark{4}, George Stoian\IEEEauthorrefmark{4}, Ye Yuan\IEEEauthorrefmark{2}  and
Guoren Wang\IEEEauthorrefmark{2}}\\
\IEEEauthorblockA{\IEEEauthorrefmark{2}School of Computer Science, Beijing Institute of Technology, China\\
\IEEEauthorrefmark{4}Department of Computing Science, University of Aberdeen, UK\\
Email: \IEEEauthorrefmark{2}luzhenyu@bit.edu.cn,
\IEEEauthorrefmark{2}yrcheng@bit.edu.cn,
\IEEEauthorrefmark{4}mingjun.zhong@abdn.ac.uk,
\IEEEauthorrefmark{4}george\_stoian@protonmail.ch,
\IEEEauthorrefmark{2}yuan-ye@bit.edu.cn,
\IEEEauthorrefmark{2}wanggrbit@126.com
}}
\begin{comment}
\end{comment}

%Note that this manuscript is submitted to IEEE Transactions on Neural Networks and Learning Systems (IEEE TNNLS)
%Special Issue on
%Causal Discovery and Causality-Inspired Machine Learning
\maketitle
\begin{abstract} 
\noindent Causal inference is to estimate the causal effect in a causal relationship when intervention is applied. Precisely, in a causal model with binary interventions, i.e., control and treatment, the causal effect is simply the difference between the factual and counterfactual. The difficulty is that the counterfactual may never been obtained which has to be estimated and so the causal effect could only be an estimate. The key challenge for estimating the counterfactual is to identify confounders which effect both outcomes and treatments. A typical approach is to formulate causal inference as a supervised learning problem and so counterfactual could be predicted. Including linear regression and deep learning models, recent machine learning methods have been adapted to causal inference. In this paper, we propose a method to estimate Causal Effect by using Variational Information Bottleneck (CEVIB). The promising point is that VIB is able to naturally distill confounding variables from the data, which enables estimating causal effect by using observational data. We have compared CEVIB to other methods by applying them to three data sets showing that our approach achieved the best performance. We also experimentally showed the robustness of our method. Source code can be found at https://github.com/kunnao/CausalVIB/tree/master.
\end{abstract}


\section{Introduction}
Causal inference \cite{gelman2006data,regressionbase,peters2017elements} is a task to estimate the effect in a causal relationship when an intervention action is made, which could be used to guide decision-making. In general, causal inference is to estimate the level of outcome changes when their causes are intervened. For example, in medical science we require to estimate the causal effect of a treatment to understand the effectiveness of the treatment applied for a particular disease. Causal inference is hard because we could only observe the factual, but would never be able to obtain the counterfactual. The key to estimate the causal effect is to estimate the counterfactual, and so the causal effect is simply the difference between factual and counterfactual. A possible approach for causal inference is the randomized controlled trails (RCT), which is a standard approach in clinical trials. In RCT, two groups of participants are randomly chosen and they are the control and treatment groups. The effect of the treatment is then computed by using the outcomes of the two groups, i.e., the average difference between their outcomes. Although RCT is considered as a most reliable approach for causal inference in practice, it could cost expensives and would be impossible in some situations. In RCT, large numbers of participants may be required to control the effect of confounding factors, which affect both the treatment and outcome. For estimating the causal effect, it is crucial to identify the confounding variables and then eliminate their effects on the outcomes so that the effect of the treatment could be correctly estimated.\\

Simply, causal inference could be formulated as a linear regression model, where the outcome is represented as the linear regression of the treatment and any other feature variables including confounding variables~\cite{gelman2006data}. The major disadvantage of linear regression approach is its limited ability to handle big data, which is the favorable circumstance in the big data era. Recent advances in deep learning techniques provide an excellent opportunity to utilise large observational data to estimate causal effects. In literature, a couple of deep neural network (DNN) approaches have been proposed for causal inference. For example, Dragonnet \cite{dragonnet} proposed an end-to-end DNN architecture with three heads to estimate treatment outcomes. The factual and counterfactual outcomes are respectively represented as two of the three heads. In this architecture, the confounding variables are supposed to be learned from the observational data, which is the main point of the proposed architecture. Variational autoencoder (VAE) is also proposed to modeling the confounding variables so that the causal effect could also be estimated using a neural network \cite{cevae}. It is interesting to note that all these methods employing deep neural networks attempt to represent the causal inference as a supervised learning problem. The important aspect for such approach is that once the causal inference is posed as a supervised learning problem, the established supervised learning algorithms could then be easily applied boosting the research in this area.\\

In this paper, we consider to use neural network inspired by variational information bottleneck~\cite{DeepVIB} to not only fit a model to the observed data, but also address hidden confounders that affects treatments or outcomes. Based on this approach, we develop a novel regularization framework that forces the model to ``forget" some hidden parameters which are not confounders and learning to extract the confounding variables from data. We present experiments in section~\ref{sec:exp} that demonstrate the advantages of this model and show that it outperforms state-of-the-art models in a variety of datasets.

\section{Causal effect models}
\label{cef}
Suppose that a dataset $D=\{X_i, T_i, Y_i\}_{i=1}^{N}$ is given, where $X$ represents the covariates of a subject, e.g., the health status; $T$ represents the treatment applied to the patient, e.g., a medication; and $Y$ represents the outcome after the treatment is applied. Note that we consider a binary variable for $T$ and so $T\in\{0,1\}$, where $T=0$ means that no medication is applied to the control group, and $T=1$ means that the medication is applied to the treatment group. For a subject $X_i$, its treatment $T_i$ can be $T_i=0$ or $T_i=1$. For a control subject $i$, since $T_i=0$ is implemented, the outcome $y_{i0}$ is called the factual outcome  which is observed; however, $T_i=1$ was never implemented, the potential outcome $y_{i1}$ is called the counterfactual outcome which is thus never able to be observed. Conversely, for a treatment subject $j$, the outcome $y_{j1}$ is the factual outcome which is observed, and $y_{j0}$ is the counterfactual outcome, which is never observed. For computing the causal treatment effect for a subject $i$, i.e., $y_{i1}-y_{i0}$, it is required to know both factual and counterfactual outcomes of a subject, but the counterfactual outcome is not obtained. Our objective is to estimate the average treatment effect (ATE) denoted by $\psi$ and individual treatment effect (ITE) when a treatment $T$, e.g., a medication, is applied to a patient.  \\

\begin{figure}[!t]
\centering
    \includegraphics[width=1.65in]{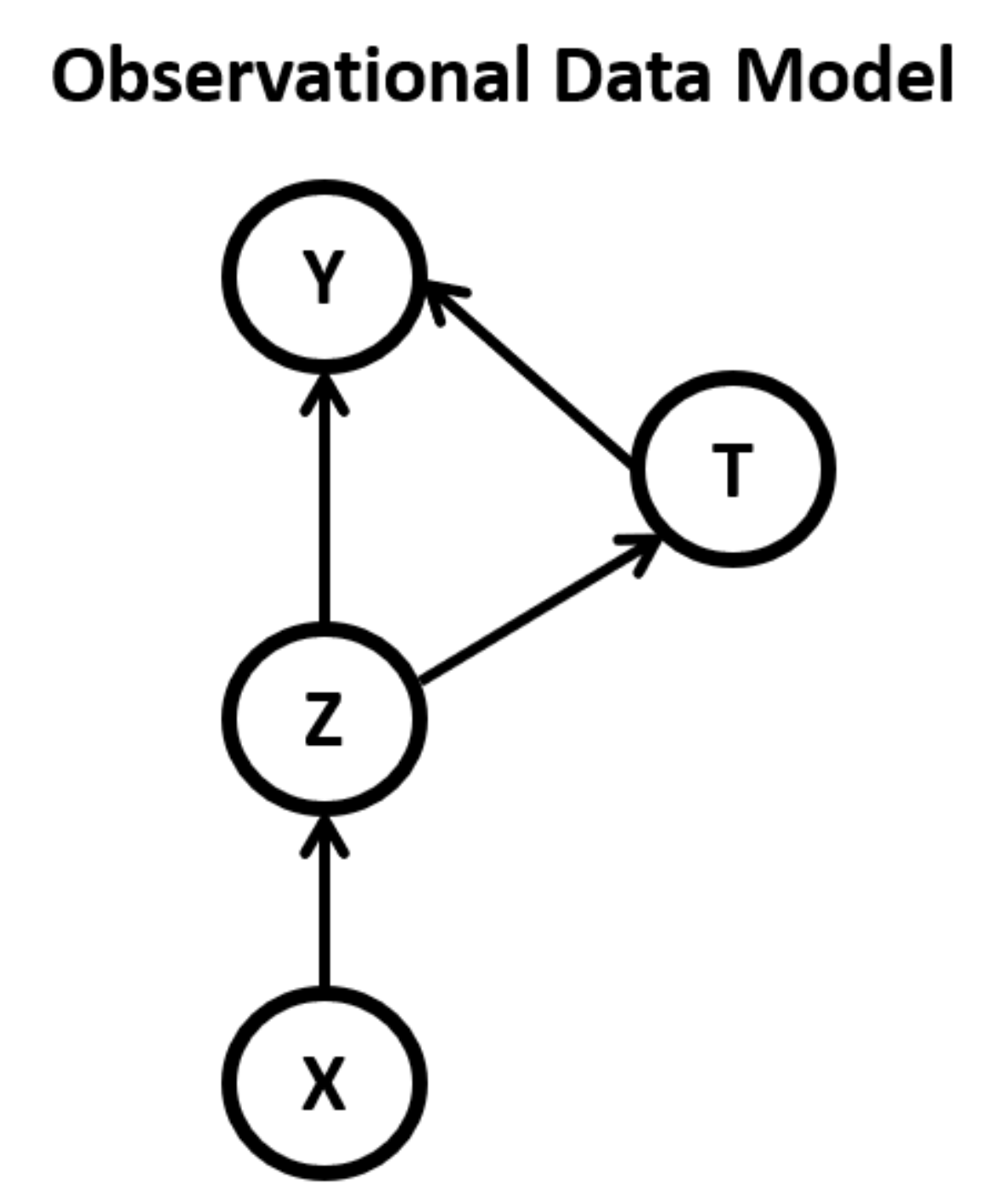} 
    \includegraphics[width=1.35in]{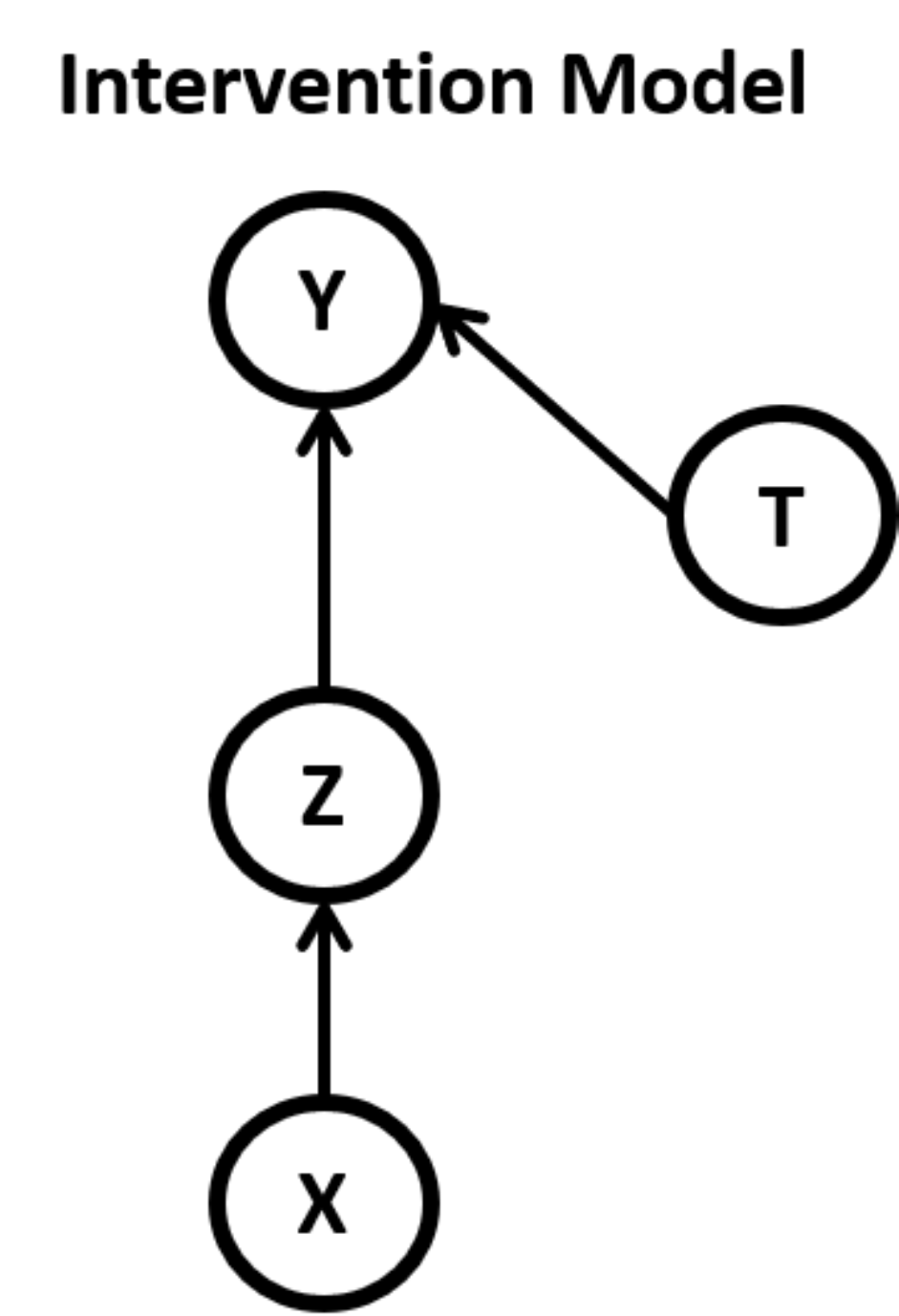}
\caption{The observation model (left) and the intervention model (right). $X$ are covariates, $Z$ are the latent confounders, $T$ is a treatment, and $Y$ is an outcome.}
\label{fig:dataMod}
\end{figure}

For a subject $i$ with a $d$-dimensional covariate ${X_i} \in {R^d}$ and its outcome being ${Y_i} \in {R^1}$ after a treatment $T_i$ is applied, we assume that the observed covariates $X_i$ include all possible causes for outcome and treatment. Those causal variables, i.e., confounders, for both treatment and outcome are hard to identify. We therefore assume the cofounders to be a collection of latent variables which could be distilled from covariates. Figure \ref{fig:dataMod} (left) represents an observational data model using this assumption, in which $X$, $T$ and $Y$ are observed, and $Z$ are the confounders. To estimate causal effect, it needs to use the intervention model (the right graph in Figure \ref{fig:dataMod}), in which the intervention treatment, i.e., Pearl's $do()$ action, is applied which is not effected by any confounder. We would also assume the following Assumptions \ref{asm:1}-\ref{asm:3}, which are sufficient to identify the treatment effect from the observed data \cite{causinf-assumption}: 

\newtheorem{assumption}{Assumption}[section]
\begin{assumption}
\label{asm:1}
The potential outcome of treatment $T$ equals the observed outcome if the actual treatment received is $T$. i.e. $\forall t\in \{0,1\},if [T=t], then [Y=Y(t)]$.\\
\end{assumption}

\begin{assumption}
\label{asm:2}
For any set of covariates $X$ the probability to receive treatment 0 or 1 is positive. i.e. $\forall X \in {R^d}:P(T=0|X) \in (0,1)\ and\ P(T=1|X) \in (0,1)$.\\
\end{assumption}

\begin{assumption}
\label{asm:3}
The potential outcomes $Y(1)$ and $Y(0)$ are independent of the treatment assignment $T$ given the covariate variables $X$, i.e. $Y \perp \!\!\! \perp T|X$.\\
\end{assumption}

Under these assumptions, the individual treatment effect could be represented as 
\begin{equation}
\begin{split}
    ITE(x_i)=&E_{Y|X,do(T_i = 1)}[Y]    - E_{Y|X,do(T_i = 0)}[Y].
\end{split}
\end{equation}
The ATE is then the expectation with respect to all the individuals and so 
\begin{eqnarray}
    \psi&=&E_X[ITE(X)]\\
    &=&E_X[E_{Y|X,do(T_i = 1)}[Y]- E_{Y|X,do(T_i = 0)}[Y]]\\
    &=&E_{Y|do(T_i = 1)}[Y]-E_{Y|do(T_i = 0)}[Y]
\end{eqnarray}
Note that Pearl's $do()$ notation is used to represent the causal relationship indicating the potential effect when the subject is intervened by applying the treatment. Note that the Assumption \ref{asm:1} is equivalent to the principle of independent mechanism (PIM) \cite{peters2017elements}, which assumes that the condional distributions in the system do not influence each other. Under the PIM assumption, the conditionals in the intervention model do not change and so are identical to those in the observational data model. This assumption allows us to apply the learned conditionals by using observation data model into the intervention model for inference.\\

\begin{figure*}[!thp]
\centering

    \includegraphics[width=0.8\textwidth]{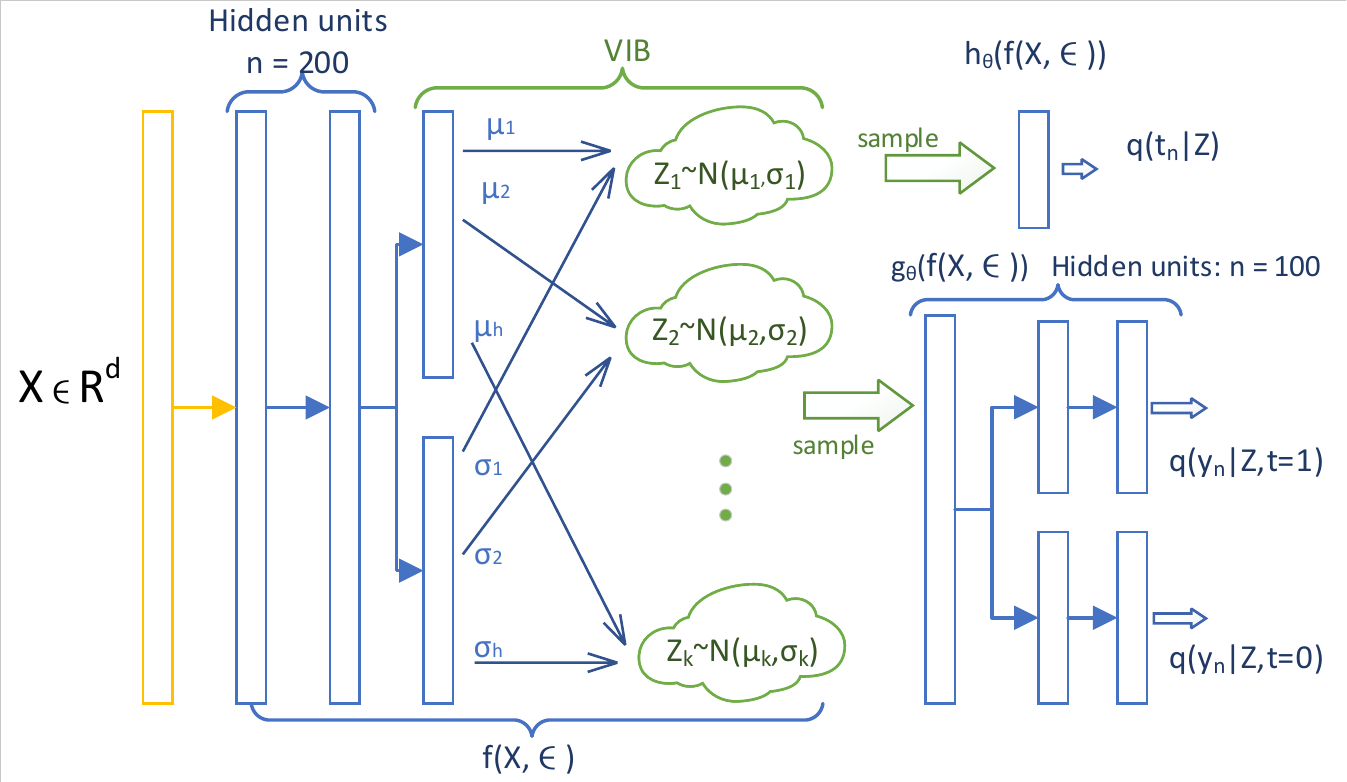}
    \caption{The architecture framework for estimating causal effect using variational information bottleneck.}
\label{fig:models}
\end{figure*}

\section{Related works}
In this section, we briefly discuss causal effect estimation methods.
Statistical methods have been proposed for causal effect estimation. For example, regression methods fit the treatment assignment and as well as the covariates to represent the outcomes~\cite{gelman2006data,regressionbase, regress2016, regress2017}. Sample re-weighting methods aim to correct the treatment assignment using observational data in order to overcome the subject selection bias. For instance, these methods include Inverse Propensity Weighting (IPW) based on propensity score~\cite{rosenbaum1987model} and confounder balancing methods~\cite{confounderbalance2017SIGKDD}. Other methods include doubly-robust approaches which combine covariate adjustment with propensity score weighting~\cite{regress2017, doubly-robost2017, debias2017}.\\

In recent years, many machine learning methods have been applied to the problem of estimating the potential outcomes and treatment effects. Most of them could be viewed as discriminative (or supervised) learning methods. For example, Bayesian Additive Regression Trees (BART)~\cite{BART2017} and Causal Forests~\cite{c-forest2018} have been used to estimate causal effect. The work in~\cite{johansson2016learning} proposed Balanced Linear Regression (BLR) and Balanced Neural Networks (BNN) to learn a balanced covariate representation for causal outcomes. Other approaches include Causal Multi-task Gaussian Processes(CMGP) where the factual and counterfactual outcomes are modelled using a vector-valued reproducing kernel Hilbert space~\cite{alaa2017bayesian} and GANITE which uses generative adversarial networks~\cite{GANITE2018} to estimate causal effect.

Methods that have similarities with our CEVIB are CEVAE~\cite{louizos2017causal} and Dragonnet~\cite{dragonnet}. All these methods have a similar neural networks structure to that of TARnet~\cite{shalit2017estimating}. An advantage of CEVIB is that this method uses variatinal bottleneck approach to represent the confounding variables as a distribution which could be able to take into account the uncertainty of the model. In the experiment section, we show that CEVIB is robust in terms of subject sample bias to support such advantage.

\section{Variational information bottleneck for causal inference}
Under the sufficient assumptions provided in Section \ref{cef}, the observed data $D=\{X, T, Y\}$ were generated from the observational data model illustrated in Figure \ref{fig:dataMod} with confounding variables $Z$. The confounders $Z$ effect both treatment $T$ and outcomes $Y$. To estimate causal effects, instead of using the obervational data model, the intervention model should be used where the treatment is intervened. However, all the conditional distributions in the intervention model is unknown, and they need to be learned from data. According to the principle of independent mechanism, the conditionals in intervention model are indentical to those in observational data model. There is hope that these conditionals could be learned from the data. According to our observational data model, we hope that all the confounding variables could be distilled from the data $X$ because $X$ contain all the possible confounders. We employ the idea of vational information bottleneck (VIB) which attempts to distill sufficient informaton from $X$ which maximumly inform both $T$ and $Y$ (see the Figure \ref{fig:dataMod}). Our VIB network architecture (CEVIB) is shown in the Figure \ref{fig:models}. Our framework is most relevant to the Dragonnet
~\cite{dragonnet}. CEVIB provides an end-to-end procedure for predicting treatment and outcome. We firstly use neural network to learn the representation $Z(x)\sim N(\mu^{k},\sigma^{k})$ following a Normal distribution, where $k$ denotes the dimension of hidden confounders. The hidden confounders are regularized by information bottleneck and predict the treatment and outcome. The regulation of variational information bottleneck enforces the model to extract hidden confounding variables as much as possible and forget those variables which are not confounders. 

\subsection{Training the observational model}
In this section, we apply the idea of VIB to train the observation model using the data. Denote $\widetilde{Y}=\{Y,T\}$. Our aim is to optimize the following problem which attempts to maximizing the mutual information between $\widetilde{Y}$ and the confounders $Z$ whist threshold the mutual information between the covariates $X$ and $Z$:
\begin{eqnarray}
\max_{\theta}I(Z,\widetilde{Y};\theta), \text{subject to}, I(X,Z;\theta)\leq I_c,
\end{eqnarray}
where $I(Z,\widetilde{Y};\theta)=\int{p(Z,\widetilde{Y})\log\frac{p(Z,\widetilde{Y})}{p(Z)p(\widetilde{Y})}d\widetilde{Y}dZ}$ represents the mutual information, and $I_c$ is a constant.
This is equivalent to maximizing the following objective function
\begin{eqnarray}
R_{IB}(\theta)=I(Z,\widetilde{Y};\theta)-\beta I(X,Z;\theta).
\end{eqnarray}

\newcounter{mytempeqncnt}
\begin{figure*}[!t]
\normalsize
\setcounter{mytempeqncnt}{\value{equation}}
\setcounter{equation}{5}
\begin{align}
%\label{lowerBound}
    L&=\frac{1}{N}\sum_{n=1}^N\left\{\int p(z|x_n)\left[t_n\log{q(y_n|z,t_n=1)}+(1-t_n)\log{q(y_n|z,t_n=0)}+\log{q(t_n|z)}-\beta \log\frac{p(z|x_n)}{r(z)}\right]dz\right\}\nonumber\\
\label{lowerBound2}
    &=\frac{1}{N}\sum_{n=1}^N\left\{\int p(z|x_n)\left[t_n\log{q(y_n|z,t_n=1)}+(1-t_n)\log{q(y_n|z,t_n=0)}+\log{q(t_n|z)}\right]dz + \beta KL[p(z|x_n)||r(z)]\right\}\nonumber\\
    &= L_1 + L_2
\end{align}

    \begin{equation}
    \label{equ:L1}
    \begin{split}
        L_1=\frac{1}{N}\sum_{n=1}^N\left\{\frac{1}{M}\sum_{m=1}^M\left[t_n\log{q(y_n|f(x_n,\epsilon_m),t_n=1)}
        +(1-t_n)\log{q(y_n|f(x_n,\epsilon_m),t_n=0)}+\log{q(t_n|f(x_n,\epsilon_m))}\right]\right\}
    \end{split}
    \end{equation}
    \begin{equation}
    \label{equ:L2}
    \begin{split}
        L_2 = \frac{\beta}{N}\sum_{n=1}^NKL[p(z|x_n)||r(z)]
    \end{split}
    \end{equation}

\setcounter{equation}{8}
\hrulefill
\vspace*{4pt}
\end{figure*}

Instead we could maximize the following lower bound
\setcounter{equation}{\value{mytempeqncnt}}
\begin{align}
    &R_{IB}(\theta)=I(Z,\widetilde{Y};\theta)-\beta I(X,Z;\theta)\nonumber\\
    &\geq\int p(x)p(\widetilde{y}|x)p(z|x)\log{q(\widetilde{y}|z)}dxd\widetilde{y}dz \nonumber\\
    &\quad- \beta \int p(x)p(z|x)\log\frac{p(z|x)}{r(z)}dxdz\nonumber\\
    &\approx\frac{1}{N}\sum_{n=1}^N\left\{\int p(z|x_n)\left[\log{q(\widetilde{y_n}|z)}-\beta \log\frac{p(z|x_n)}{r(z)}\right]dz\right\}\nonumber\\
    &=L.
\end{align}

In the causal effect model (see the observation model in Figure \ref{fig:dataMod}), our outputs given the latent variable has the form
\[
q(\widetilde{y}|z)=q(y|z,t=1)^{t}q(y|z,t=0)^{1-t}q(t|z)
\]
So the lower bound can be represented as the equation~(\ref{lowerBound2}).

Suppose the latent variable $z$ has dimension $K$, we assume a diagonal multivariate Normal distribution,
\begin{equation}
\nonumber
    p(z|x_n)=\prod_{k=1}^Kp(z_k|x_n)=\prod_{k=1}^KN(z_k|\mu_{k}(x_n),\sigma^2_k(x_n))
\end{equation}

where $\mu_k$ and $\sigma_k$ are neural networks. Therefore, we can use change of variables to draw a sample $z_k=f_k(x_n,\epsilon)=\mu_{k}(x_n)+\sigma_k(x_n)\epsilon$ where $\epsilon\sim N(0,1)$. Denote $f(x_n,\epsilon)=[f_1(x_n,\epsilon),\cdots,f_K(x_n,\epsilon)]^T$. So $L1$ can be expressed as equation~(\ref{equ:L1}).
We define the following distributions for outcome and treatment variables
\setcounter{equation}{8}
\begin{align}
    \log{q(y_n|f(x_n,\epsilon_m),t_n)}\approx-(y_n-g^{t_n}_{\theta}(f(x_n,\epsilon)))^2\\
    \log{q(t_n|f(x_n,\epsilon_m))}\approx-(t_n-h_{\theta}(f(x_n,\epsilon)))^2
\end{align}

where $g^{t_n}_\theta(.)$ and $h_\theta(.)$ are neural networks. Therefore, our task is then to learn the neural networks $g^{t_n}_\theta(.)$, $h_\theta(.)$, $\mu_k(.)$ and $\sigma_k(.)$.
For the $L_2$, we use the identity
\begin{equation}
\nonumber
\begin{split}
    KL&\left(N\left((\mu_1,\cdots,\mu_K)^T,diag(\sigma_1^2,\cdots,\sigma_K^2)\right)||N(0,1)\right)\\
    &=\frac{1}{2}\sum_{k=1}^K(\sigma_k^2+\mu_k^2-1-ln(\sigma_k^2))   
\end{split}
\end{equation}
So we have
\begin{equation}
\nonumber
\begin{split}
L_2 &= \frac{\beta}{N}\sum_{n=1}^NKL[p(z|x_n)||r(z)]\\
&=\frac{\beta}{N}\sum_{n=1}^N\left[\frac{1}{2}\sum_{k=1}^K(\sigma^2_k(x_n)+\mu_{k}(x_n)-1-ln(\sigma^2_k(x_n)))\right]
\end{split}
\end{equation}
Our CEVIB is to train to maximize $L_1$ and $L_2$ using observational data, and so all the conditional distributions are trained, which can then be used in the intervention model to estimate causal effects, which we will describe in the following subsection.\\

According to the {\em Assumption 2.2}, the treatment outcome $Y$ and the treatment $T$ are independent given the confounder variables $Z$. This could be achieved by minimizing the following mutual information between $Y$ and $T$ given $Z$:
\begin{eqnarray}
    I(Y,T|X)&=&\int\int\int  p(X)p(Y,T|X)\log{\frac{p(Y,T|X)}{p(Y|X)p(T|X)}}dYdTdX\\
    &=&\int\int\int  p(X,Y,T)\left[\log{p(Y,T|X)}-\log{p(Y|X)} -\log{p(T|X)} \right]dYdTdX\\
    &=&\int\int\int  p(X,Y,T)\left[\log{p(Y,T|X)}-\log{\int p(Y|Z)p(Z|X)dZ} -\log{\int p(T|Z)p(Z|X)dZ} \right]dYdTdX\\
    &\leq& \int\int\int  p(X,Y,T)\left[\log{p(Y,T|X)}-{\int \log{p(Y|Z)}p(Z|X)dZ} -{\int \log{p(T|Z)}p(Z|X)dZ} \right]dYdTdX\\
    &\propto&-\frac{1}{N}\sum_{n=1}^N\int p(Z|x_n)\left[\log{p(y_n|Z)}+\log{p(t_n|Z)}\right]dZ
\end{eqnarray}
Note that the inequality comes from Jensen's inequality. We can see that the VIB lower bound automatically satisfies the conditional independent assumption.

\subsection{Estimating causal effects}
After training, the average treatment affect $\tau$ can be calculated using,
\begin{eqnarray}
    \tau = \int Yp(Y|do(T=1))dY - \int Yp(Y|do(T=0))dY
\end{eqnarray}
We can then apply the $do-calculus$ according to the intervention model structure (see the intervention model in Figure \ref{fig:dataMod})
\begin{align}
    &p(Y|do(T=t))\nonumber\\
    &=\int p_{do(T=t)}(Y,Z,X,T=t)dZdX\\
    &=\int p_{do(T=t)}(Y|Z,X,T=t)p_{do(T=t)}(Z,X,T=t)dZdX\\
    &=\int p_{do(T=t)}(Y|Z,T=t)p(Z|X)p(X)p_{do(T=t)}(T=t)dZdX\\
    &=\int p(Y|Z,T=t)p(Z|X)p(X)dZdX\\
    &\approx \int q(Y|Z,T=t)p(Z|X)p(X)dZdX
    %&\approx& \frac{1}{N}\sum_{n=1}^N\left[\frac{1}{M}\sum_{m=1}^Mp(Y|z_m,T=t; x_n)\right]
\end{align}
where we have applied the approximation $p(Y|Z,T=t) \approx q(Y|Z,T=t)$. Note that all these conditionals involving computing $p(Y|do(T=t))$ have been trained using observational model.
We then draw samples from $p(Y|do(T=t))$ which will be used to calculate the average treatment affect. To draw samples from $p(Y|do(T=t))$, we do the following process
\begin{itemize}
    \item $x^{(i)}\sim p(X)$, i.e., draw a sample $x^{(i)}$ from $p(X)$
    \item $z^{(i)}\sim p(Z|x^{(i)})$
    \item $y_t^{(i)}\sim q(Y|z^{(i)},T=t)$
\end{itemize}
This provides a Monte Carlo estimate for the average treatment affect,
\[
\hat{\tau}=\frac{1}{N_1}\sum_{i=1}^{N_1}y_1^{(i)} - \frac{1}{N_0}\sum_{i=1}^{N_0}y_0^{(i)}
\]
We will use this method to estimate causal effects in the following experiments.

\section{Experiments}
\label{sec:exp}
\subsection{The data sets and training setup}
Due to the difficulty of gathering treatment effects factuals for both control and treatment, evaluating the methods for estimating causal effects may have to use synthetic or semi-synthetic data sets. Our experiments will apply CEVIB to three semi-synthetic benchmark data sets: \textbf{IHDP}~\cite{yao2018representation}, \textbf{Twins}~\cite{GANITE2018} and \textbf{ACIC}~\cite{2018Benchmarking}. \\

\textbf{IHDP}: This data is constructed from the Infant Health and Development Program (IHDP). There are 100 files in which each file contains 747 subjects\footnote{The IHDP data set is available at https://github.com/Osier-Yi/SITE/tree/master/data}. Both factual and counterfactual are provided for each subject, which provides a ground truth for evaluating causal inference algorithms. \\
\begin{comment}
We use 1000 realizations from the NPCI package~\cite{Dorie2016} which has 747 observed values.
\end{comment}

\textbf{Twins}: This data is a benchmark task that utilizes data from twin births in the USA between 1989-1991. There are 11399 subjects in this data\footnote{The Twins data set is available at https://github.com/jsyoon0823/GANITE/tree/master/data or https://github.com/AMLab-Amsterdam/CEVAE/tree/master/ datasets/TWINS.}. The samples in the data set are all twins, $t = 1$ represents the heavier baby, and the outcome $Y$ corresponds to the mortality of each of the twins in their first year of life. \\

\textbf{ACIC}: This data is a collection of semi-synthetic datasets derived from the linked birth and infant death data (LBIDD)~\cite{2018Benchmarking}. It was developed for the 2018 Atlantic Causal Inference Conference competition (ACIC)\footnote{We use the scaling folder in ACIC to evaluate our methods. The ACIC data set is available at https://github.com/IBM-HRL-MLHLS/IBM-Causal-Inference-Benchmarking-Framework/tree/master/data/LBIDD.}, which include 30 different data generating process settings with subject sample sizes from 1,000 to 50,000.\\

To train CEVIB, we use the architecture in Figure \ref{fig:models}. For both IHDP and ACIC experiments, we randomly split each file data into test/validation/train with proportion 63/27/10 and report the In Sample\footnote{In Sample uses all observational data for both training and prediction.} and Out of Sample\footnote{Out of Sample uses the training set to train and the test set for prediction.} errors, and repeat the procedure for 25 times. For Twins experiments, we randomly split the data into test/validation/train with proportion 56/24/20 and report the In Sample and Out of Sample errors, and repeat the procedure for 50 times. \\

To evaluate the performance, we report the results of the following metrics for each data set, which are the absolute error in average treatment effect  
\[{\epsilon_{\rm{ATE}}}{\rm{ = }}|\frac{1}{N}\sum\limits_{i = 1}^N {(y_1^{(i)} - y_0^{(i)})} - \frac{1}{N}\sum\limits_{i = 1}^n {(\hat y_1^{(i)} - \hat y_0^{(i)}})|, \] the Precision in Estimation of Heterogeneous Effect (PEHE)
\[{\epsilon _{\rm{PEHE}}}{\rm{ = }}\frac{1}{N}\sum\limits_{i = 1}^N ({(y_1^{(i)} - y_0^{(i)}) - ({{\hat y_1^{(i)} - \hat y_0^{(i)}))}^2}},\]
and the relative PEHE:
\[
PEHE_{REL} = \frac{1}{N}\sum_{i=1}^N\left(\frac{(y_1^{(i)}-y_0^{(i)})-(\hat{y}_1^{(i)}-\hat{y}_0^{(i)})}{(y_1^{(i)}-y_0^{(i)})}\right)^2.
\]
Note that the relative PEHE could be able to remove the scaling effect on PEHE.\\

We compared CEVIB with Dragonnet~\cite{dragonnet}, the regularized Dragonnet (Dragon-tarreg) \cite{dragonnet} and causal effect variational auto-encoder(CEVAE)~\cite{cevae}. We also compare to Dragonnet-tarreg which is a Dragonnet using a targeted regularization method based on the augmented inverse probability weighted(AIPW) estimator. In addition, we also compared CEVIB with least squares regression using treatment as a feature (OLS/LR1), separate least squares regressions for each treatment (OLS/LR2), BLR~\cite{LearningFromCF2016}, BART~\cite{BART2017}, CForest~\cite{c-forest2018}, BNN~\cite{LearningFromCF2016}, $CFR_{wass}\ and \  TARNet$~\cite{shalit2017estimating}, and GANITE~\cite{GANITE2018}.

\subsection{Results}
\begin{figure*}[h]

\centering
\subfigure[${IHDP}$]     {
    \includegraphics[width=0.45\textwidth]{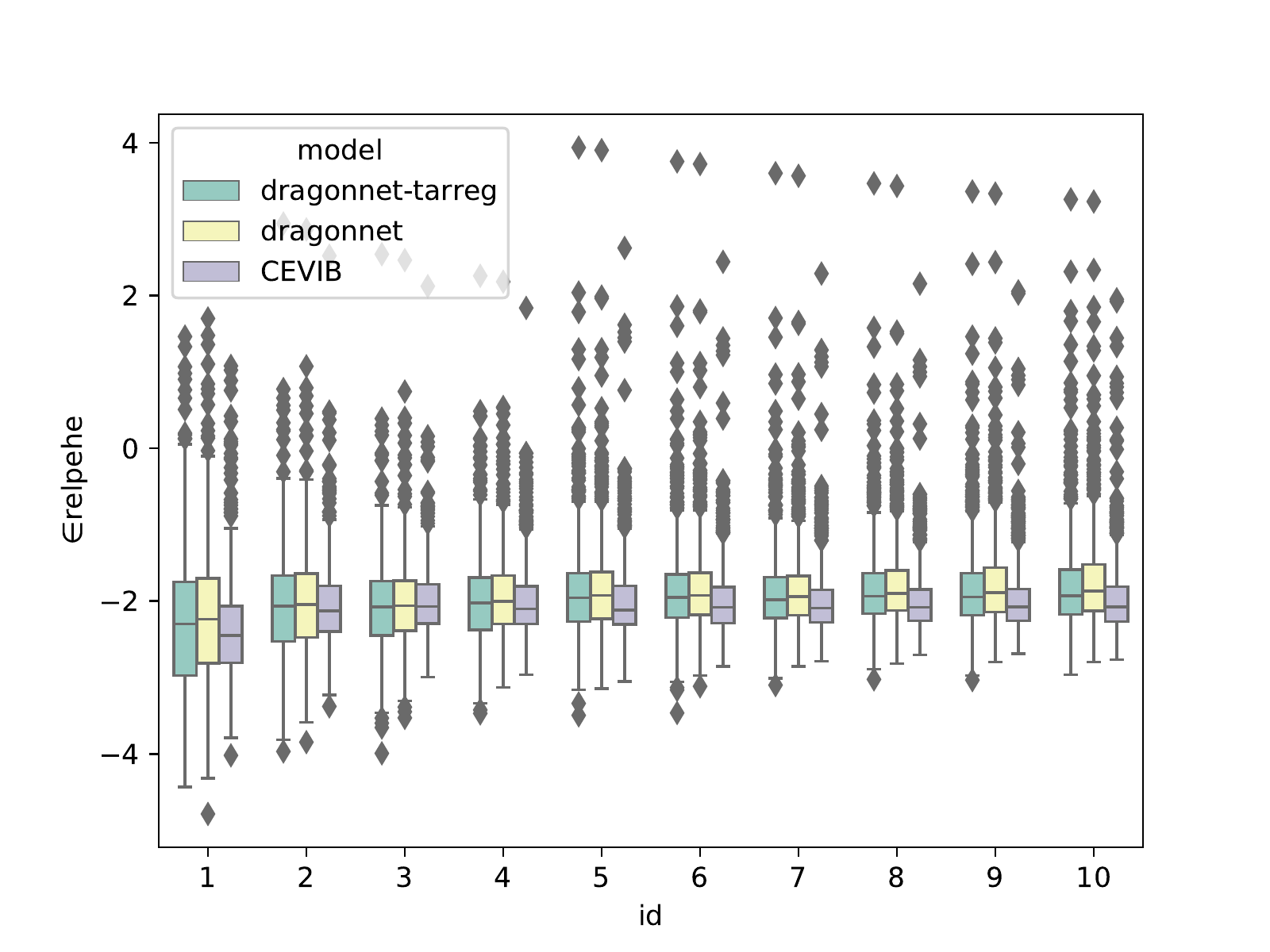}
    }
\subfigure[${ACIC}$]     {    
    \includegraphics[width=0.45\textwidth]{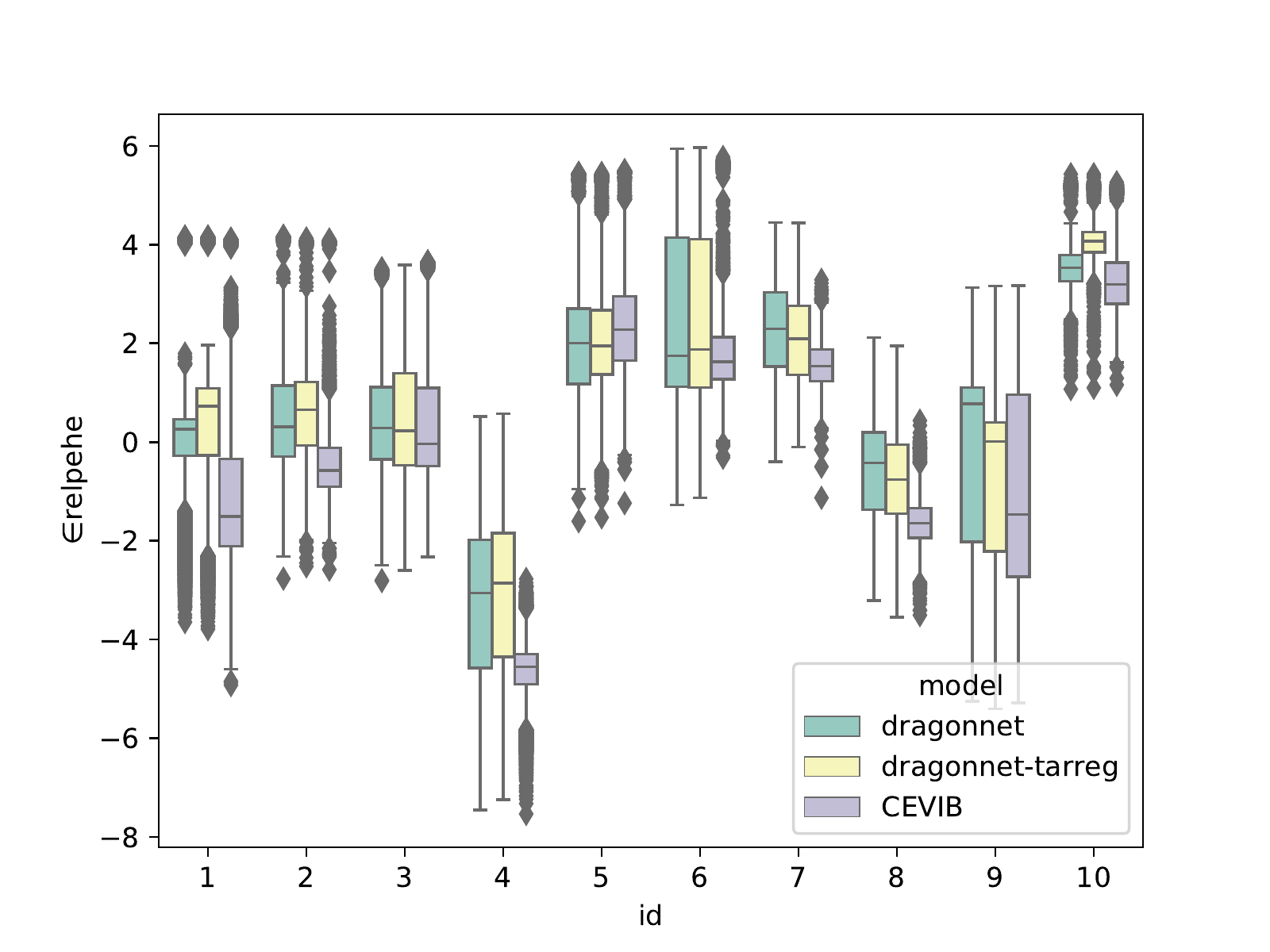}
}
\caption{The result of relative PEHE (logarithm) when CEVIB, Dragonnet, and Dragonnet-tarreg were applied to IHDP and ACIC. Ten data files were randomly chosen to plot their distributions using the replicated results.}
\label{fig:relpehe_res}
\end{figure*}

For comparison purpose, various methods described in the previous subsection are applied to the data sets IHDP and Twins. For the error metrics ATE and PEHE, both the within samples and out of samples were computed. The results are shown in Table~\ref{tab:expriment} which shows that our CEVIB outperforms all of other methods on the data IHDP. On the Twins data set, CEVIB is among the best results. Due to computational limits, on the ACIC data, we compared CEVIB to CEVAE, Dragonnet, and Dragon-tarreg. The results are shown in the Table \ref{tab:expriment2}. The results clearly indicate that our CEVIB outperforms all the other methods across all the error metrics. To have a more clearer comparison, we randomly chose 10 data files from IHDP and ACIC to plot the distributions for the replicated results using CEVIB, Dragonnet, and Dragonnet-tarreg. The results are plotted in the Figure \ref{fig:relpehe_res}. We showed the logarithmic values for relative PEHE. It shows that accross all the data files, CEVIB outperformed the other methods.

\begin{table*}[!tbp]
\caption{The results of applying various models to IHDP and Twins data. Best results are denoted in bold. Note that $within-s$ means that training data were used for both training and prediction and $out-of-s$ means that the models were trained on training data and tested on the test data.}
\label{tab:expriment}
\centering
\resizebox{\textwidth}{!}{
\begin{tabular}{@{}lllllllll@{}}
\toprule
\hline
& \multicolumn{8}{c}{Datasets(Mean +- std)}        \\ \hline
& \multicolumn{4}{c|}{IHDP} 
& \multicolumn{4}{c|}{Twins} \\ \hline
Method                  
& $\sqrt{\epsilon_{PEHE}^{within-s}}$ & $\epsilon_{ATE}^{within-s}$
& $\sqrt{\epsilon_{PEHE}^{out-of-s}}$ & $\epsilon_{ATE}^{out-of-s}$     
& $\sqrt{\epsilon_{PEHE}^{within-s}}$ & $\epsilon_{ATE}^{within-s}$
& $\sqrt{\epsilon_{PEHE}^{out-of-s}}$ & $\epsilon_{ATE}^{out-of-s}$     \\ \hline
OLS-1               & 5.8±.3  & .73±.04 & 5.8±.3  & .94±.06 & .319±.001 & .0038±.0025 & .318±.007 & .0069±.0056 \\
OLS-2               & 2.4±.1  & .14±.01 & 2.5±.1  & .31±.02 & .320±.002 & .0039±.0025 & .320±.003 & .0070±.0059 \\
BLR                 & 5.8±.3  & .72±.04 & 5.8±.3  & .93±.05 & .312±.003 & .0057±.0036 & .323±.018 & .0334±.0092 \\
k-NN                & 2.1±.1  & .14±.01 & 4.1±.2  & .79±.05 & .333±.001 & .0028±.0021 & .345±.007 & \textbf{.0051±.0039} \\
BART                & 2.1±.1  & .23±.01 & 2.3±.1  & .34±.02 & .347±.009 & .1206±.0236 & .338±.016 & .1265±.0234 \\
RF                  & 4.2±.2  & .73±.05 & 6.6±.3  & .96±.06 & .306±.002 & .0049±.0034 & .321±.005 & .0080±.0051 \\
CF                  & 3.8±.2  & .18±.01 & 3.8±.2  & .40±.03 & .366±.003 & .0286±.0035 & .316±.011 & .0335±.0083 \\
BNN                 & 2.2±.1  & .37±.03 & 2.1±.1  & .42±.03 & .325±.003 & .0056±.0032 & .321±.018 & .0203±.0071 \\
CFRW                & \textbf{.71±.0}  & .25±.01 & \textbf{.76±.0}  & .27±.01 & .315±.007 & .0112±.0016 & .313±.008 & .0284±.0032 \\
CEVAE               & 2.7±.1  & .34±.01 & 2.6±.1  & .46±.02 & .341±.006 & .0065±.0040 & .373±.012 & .0679±.0212 \\
TARNet              & .88±.0  & .26±.01 & .95±.0  & .28±.01 & .317±.005 & .0108±.0017 & .315±.003 & .0151±.0018 \\
GANITE              & 1.9±.4  & .43±.05 & 2.4±.4  & .49±.05 & \textbf{.289±.005} & .0058±.0017 & \textbf{.297±.016} & .0089±.0075 \\
Dragonnet           & 1.31±.4 & .14±.02 & 1.32±.5 & .21±.04 & .322±.001 & .0092±.0078 & .317±.001 & .0074±.0092 \\
Dragon-tarreg       & 1.22±.3 & .14±.01 & 1.30±.3 & .20±.05 & .322±.0017 & .0060±.0088 & .318±.002 & .0060±.0101 \\
%CEVIB             & 1.32±.1 & .11±.01 & 1.30±.6 & .18±.02 & .320±.0003 & .0016±.0009 & .315±.001 & .0055±.0018 \\
CEVIB               & .85±.1  & \textbf{.12±.01} & .92±.2  & \textbf{.15±.02} & .320±.0003 & \textbf{.0016±.0009} & .315±.001 & .0055±.0018 \\ \hline
\end{tabular}}
\end{table*}

\begin{table*}[!tbp]
\caption{The results for the methods applied to ACIC data.}
\label{tab:expriment2}
\centering
\begin{tabular}{@{}lllll@{}}
\toprule
\hline
& \multicolumn{4}{c}{ACIC(Mean +- std)}        \\ \hline
Method                  
& $\sqrt{\epsilon_{PEHE}^{within-s}}$ & $\epsilon_{ATE}^{within-s}$
& $\sqrt{\epsilon_{PEHE}^{out-of-s}}$ & $\epsilon_{ATE}^{out-of-s}$     \\ \hline
CEVAE                   & 131±1.8  & 50.2±1.3 & 118±2.2  & 51.9±1.4 \\
Dragonnet               & 67.8±2.0 & 25.8±2.1 & 68.2±2.6 & 26.3±2.1 \\
Dragon-tarreg           & 62.5±2.0 & 24.8±3.1 & 62.2±2.4 & 24.5±3.0 \\
CEVIB                   & \textbf{43.9±0.8} & \textbf{14.2±1.7} & \textbf{44.3±1.8} & \textbf{14.7±1.9} \\ \hline
\end{tabular}
\end{table*}

\subsection{Robustness}
\label{subsec:simu}
\begin{figure*}[h]

\centering
\subfigure[$\epsilon_{ATE}^{out-of-s}$]     {
    \includegraphics[width=0.45\textwidth]{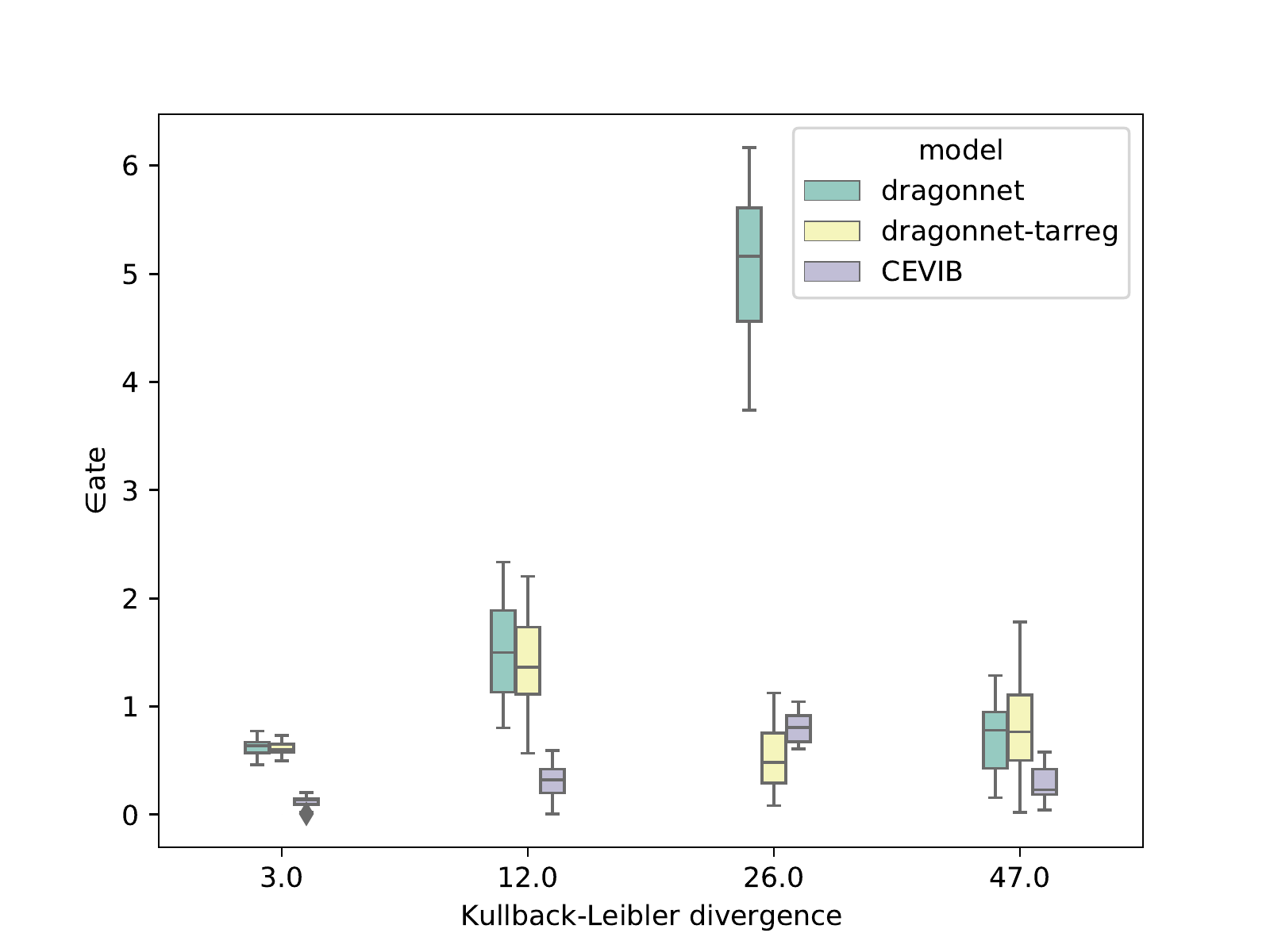}
    }
\subfigure[$\sqrt{\epsilon_{PEHE}^{out-of-s}}$]     {    
    \includegraphics[width=0.45\textwidth]{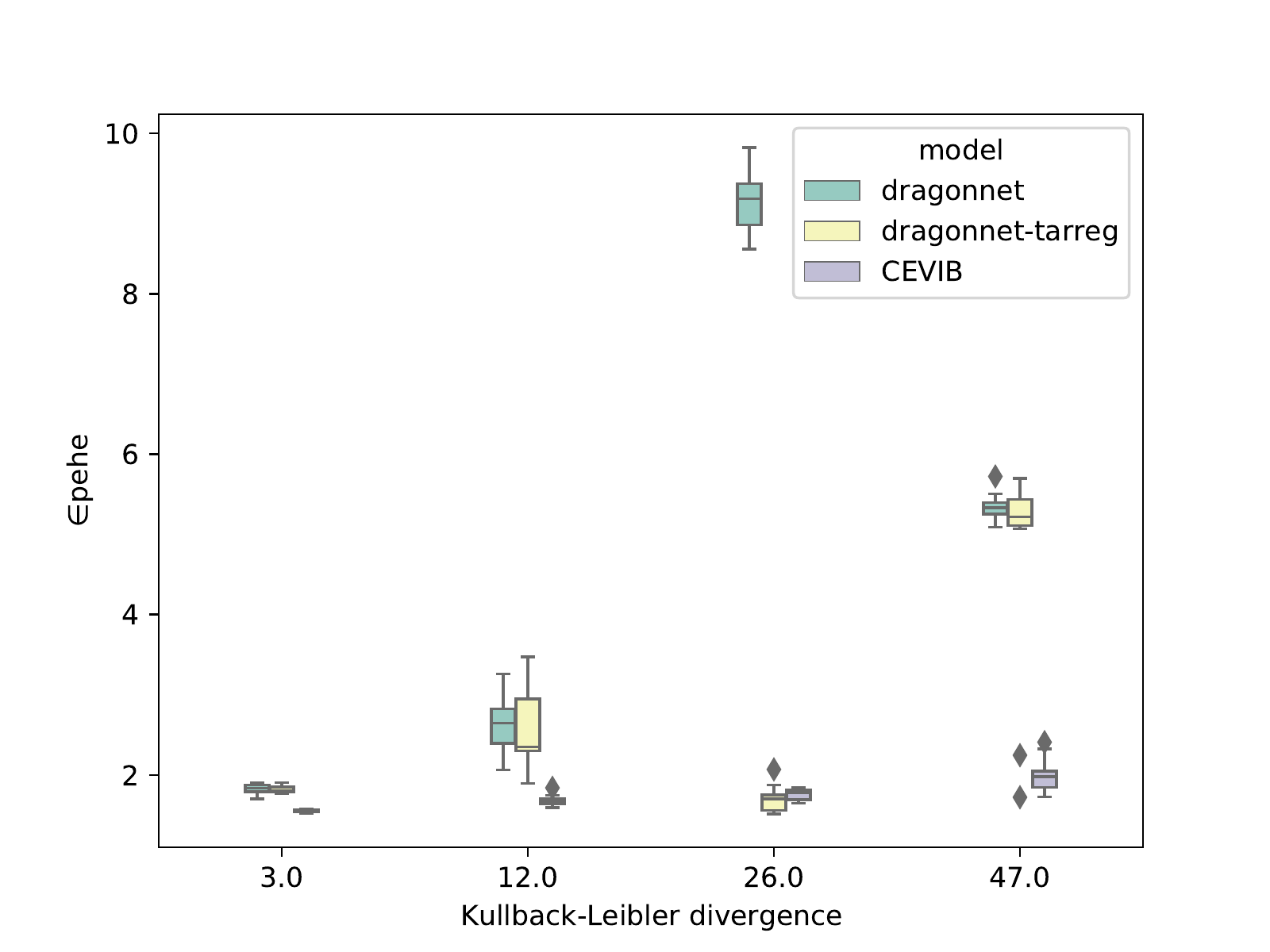}
}
\caption{The sensitivity of CEVIB, Dragonnet, and Dragonnet-tarreg to the subject sample bias. The larger KL divergence indicates more difference between the distributions of the control and the treatment samples. }
\label{fig:simu_res}
\end{figure*}

We also evaluated the robustness of CEVIB on a synthetic data set via varying selection bias. We assume the control and treatment subjects are from two different distributions. The distance between the two distributions can be evaluated by using Kullback-Leibler (KL) divergence. If the KL distance is larger, and so is the selection bias. Similar to~\cite{yao2018representation}, the synthetic data were generated with the following procedure. 5,000 control subject samples were randomly generated from $N (0^{10\times1}, 0.5 × (\Sigma+{\Sigma}^{T}))$ and 2500 treatment samples from $N (\mu^{10\times1}, 0.5 × (\Sigma+{\Sigma}^{T}))$, where $\Sigma$ was generated using Uniform distributions $U((-1,1)^{10\times10})$. By changing the value of $\mu$, data with different selection bias can be generated. The outcome is generated by $Y=W^{T}X+\epsilon$, where $W \sim U((-1,1)^{10\times2})$ and $\epsilon\sim N(0^{2\times1}, 0.1\times I^{2\times 2})$.
In this experiment, we compared CEVIB against Dragonnet and Dragonnet-tarreg. We set $\mu \in [1,2,3,4]$, all the algorithms ran 50 replications using the generated data. The results are shown in figure~\ref{fig:simu_res}. We observed that when KL values get larger, the errors are also larger.
It shows that Dragonnet was more sensitive to the bias comparing to CEVIB and Dragonnet-tarreg. For both error metrics, it shows that both CEVIB and Dragonnet-tarreg were similar in terms of sensitivity with respect to the bias. This may indicate that CEVIB and Dragonnet-tarreg could have a better ability to extract the latent confounders from observational data.

\begin{comment}

% \section{Tool Kit For Causal Effect Estimating from Observational Data}
% In recent years, the intersection of causal inference and machine learning has become an active area of research. In order to help causal effect research and apply these methods for a wider audience, We develop a Python package called Causal Effect Tool Kit(CETK) that provides causal inference methods based on machine learning algorithms. CETK enables users to estimate the Average Treatment Effect (ATE) and Conditional Average Treatment Effect (CATE), which are essentially the effect at the social or individual level. What is more, users can train the provided model by their own data, propose their novel methods, and evaluate the methods on series datasets.
% The framework of thefirst version CETK is shown in figure~\ref{fig:cetk}.

% \begin{figure*}[h]
% \label{fig:cetk}
% \centering
% \subfigure[]     {
%     \includegraphics[width=1\textwidth]{images/CETK.pdf}
%     }
% \caption{The result of simulation experiment}
% \end{figure*}
% \textbf{Future Development.} We plan to enable CETK to be used for more types of causal inference tasks, e.g. causal inference with multiple causes, continuous treatment effects or Non-i.i.d data. We welcome everyone to try out CETK on different causal methods and contribute to the development of this package.

\end{comment}

\section{Conclusion}
In this paper, we have proposed a variational information bottleneck (VIB) approach to estimate causal effects. The interesting point of using VIB is that VIB is able to distill compact information representing the latent confounders from data. Representing confounding variables is important because the effect of confounding variables on outcome can then be integrated out so that the effect of treatment could be purified. We showed that causal inference could be represented as a prediction problem. We have used VIB to train a model to predict both factual and counterfactual outcomes and so the causal effect can be calculated. The proposed algorithm CEVIB was applied to three data sets and compared to other methods showing that our method outperformed other approaches. We also demonstrated the robustness of our method.

%\bibliographystyle{johd}
%\bibliography{bib}
\bibliographystyle{abbrv}
\bibliography{bib.bib}

\end{document}